\documentclass[10pt,twocolumn,letterpaper]{article}
\usepackage[pagenumbers]{cvpr}

\usepackage{xcolor} 
\usepackage{soul}
\usepackage{pifont} 
\usepackage{graphicx}
\usepackage{multirow}
\usepackage{tabularx}
\usepackage{enumitem}
\usepackage{amsfonts}
\usepackage{booktabs}
\usepackage{adjustbox}
\usepackage{pifont}
\usepackage{subcaption}
\usepackage{comment}
\usepackage{algorithm}
\usepackage{algpseudocode}
\usepackage{diagbox}
\usepackage{gensymb}
\usepackage{multicol}
\usepackage{amssymb}
\usepackage[
  separate-uncertainty = true,
  multi-part-units = repeat
]{siunitx}

\usepackage[pagebackref,breaklinks,colorlinks]{hyperref}

\usepackage{makecell}

\usepackage{longtable}

\newcommand{\appropto}{\mathrel{\vcenter{
  \offinterlineskip\halign{\hfil$##$\cr
    \propto\cr\noalign{\kern2pt}\sim\cr\noalign{\kern-2pt}}}}}

\makeatletter
\AtBeginDocument{\let\hl\@firstofone}
\makeatother

\begin{document}

\title{Lebanon Solar Rooftop Potential Assessment using\\ Buildings Segmentation from Aerial Images}

\author{Hasan~Nasrallah, Abed Ellatif~Samhat
\thanks{H. Nasrallah and A. Samhat are with the Lebanese University.}\\
\and
Yilei~Shi, Xiaoxiang Zhu
\thanks{Y. Shi and X. Zhu are with the Technical University of Munich.}\\
\and
Ghaleb Faour and Ali J. Ghandour
\thanks{G. Faour and A.J. Ghandour are with the National Center for Remote Sensing - CNRS, Lebanon. Corresponding Author Email: aghandour@cnrs.edu.lb}\\
}

\maketitle

\begin{abstract}
Estimating solar rooftop potential at a national level is a fundamental building block for every country to utilize solar power efficiently. Solar rooftop potential assessment relies on several features such as building geometry, location, and \hl{surrounding facilities}. Hence, national-level approximations that do not take these factors into deep consideration are often inaccurate. This paper introduces Lebanon's first comprehensive footprint and solar rooftop potential maps using deep learning-based instance segmentation to extract buildings' footprints from satellite images. A photovoltaic panels placement algorithm that considers the morphology of each roof is proposed. \hl{We show that the average rooftop's solar potential can fulfill the yearly electric needs of a single-family residence while using only 5\% of the roof surface. The usage of 50\% of a residential apartment rooftop area would achieve energy security for up to 8 households.} We also compute the average and total solar rooftop potential per district to localize regions corresponding to the highest and lowest solar rooftop potential yield. \hl{Factors such as size, ground coverage ratio and $PV_{out}$ are carefully investigated for each district. Baalbeck district yielded the highest total solar rooftop potential despite its low built-up area. While, Beirut capital city has the highest average solar rooftop potential due to its extremely populated urban nature.} Reported results and analysis reveal solar rooftop potential urban patterns and provides policymakers and key stakeholders with tangible insights. \hl{Lebanon's total solar rooftop potential is about 28.1 $TWh/year$, two times larger than the national energy consumption in 2019.}
\end{abstract}

\section{Introduction}

Building footprints extraction from aerial imagery is essential for many urban applications, including geographical databases, land use, and change analysis. Fully automated extraction and recognition of buildings' footprints geometries can help estimate the solar potential of every rooftop. Estimating rooftops' solar potential provides more insight into how a given country can efficiently utilize renewable resources for solar power generation.

Solar power harvesting paves the way for ensuring a greener future and a better economic status. The first step to accomplish this task is acquiring rooftop geometries that we tackle via satellite imagery analysis and object segmentation. 

Segmentation of urban aerial imagery is currently undergoing significant attention in the research community and notable development efforts in the industry. Remote sensing images are usually complex and characterized by significant intra-class variations, and often low inter-class variations \cite{ALSHEHHI2017139}. 

Deep learning significantly reduces the time and cost required for aerial imagery segmentation due to its capabilities in automatically extracting features and patterns present in large scenes. This work uses deep convolutional neural networks to extract rooftop geometries and obtain Lebanon's first complete and comprehensive urban map. 

We then use a solar panel placement algorithm that considers the area and morphology of every rooftop to estimate the number of panels that can fit on top. Automated solar rooftop potential estimation is made feasible given the corresponding photovoltaic (PV) power generated per unit of the installed panels.

The contribution of this paper is three-folds: \textit{(i)} design the first complete Lebanese buildings' footprints map using instance segmentation from satellite images, \textit{(ii)} present also the first Lebanese solar rooftop potential map by estimating the solar potential of every rooftop morphology, \textit{(iii)} thorough analysis and insights of solar rooftop potential trends for each Lebanese district.

The rest of the paper is organized as follows: Section \ref{review} shows the literature review related to the use of deep learning models for building footprints detection and solar estimation from aerial images. We present in Section \ref{studyarea} the TUM dataset created for the scope of segmenting Lebanese buildings. Buildings footprints map is discussed in Section \ref{method}. Solar rooftop potential assessment approach is presented in Section \ref{solar} and associated results are revealed in Section \ref{resuklts}. Finally, Section \ref{conclusion} concludes the paper.

\section{Related Work}
\label{review}

\subsection{\hl{Buildings' Segmentation}}
Several techniques tackle the problem of buildings segmentation from aerial imagery. In \cite{GGCNN}, the authors use gated graph convolutional neural networks to output a Truncated Signed Distance Map (TSDM), which is then converted into a semantic segmentation mask of buildings. In \cite{RA-FCN}, the authors propose two plug-and-play modules to generate spatial augmented, and channel augmented features for semantic segmentation from satellite images. In \cite{Sat+GIS}, the authors use augmentations like slicing, rescaling, and rotations, in addition to GIS data, to improve buildings' footprint extraction. However, a more direct approach is presented in \cite{ternausv2} where the authors only use a semantic segmentation network with an additional output mask designating spacing between nearby buildings to separate building instances.\\
\hl{Similar to \mbox{\cite{ternausv2}} approach, we adopted in this work a multi-class semantic UNet-like architecture, followed by a watershed post-processing step to extract buildings' instances. However, in our proposed method, we introduced a third class which is buildings' borders to separate nearby instances. Indeed, employing an additional buildings' borders class improved F-score considerably with more than 10\%. Specific details and discussions about this third class are beyond the scope of this paper.}
\subsection{\hl{Solar Rooftop Potential Estimation}}
Solar Rooftop Potential Estimation is currently drawing the attention of geospatial deep learning researchers due to its effectiveness in accurately predicting the potential usage of rooftops for solar power generation. Significant advances in this field have been made using statistical models, computer vision, numerical analysis, and geographic information systems. In \cite{case_bei}, the authors use collected data about the solar radiation and the rooftop areas in Beirut city to estimate solar rooftop potential, without taking into consideration each rooftop morphology. A similar approach was conducted in \cite{solar_india} for a city in India, however the total building rooftop area in the city was estimated based on the building footprint extraction from satellite imagery using deep convolutional networks. In \cite{solar_uav}, authors use high precision photographic sensors mounted on a UAV to scan and create a digital surface model (DSM) for a single building. Further statistical analysis of the output DSM model, including shading analysis, solar irradiance estimation, and panel placement, was conducted to estimate a single roof's solar potential.  In \cite{solar_ref1}, the authors use satellite imagery to divide and segment building rooftops into sections using convolutional neural networks. They predict each section's pitch, azimuth, and shading mask and then use a greedy algorithm to place solar panels on rooftops to estimate their solar potential. A similar approach was adopted in \cite{solar_china} to estimate the solar potential at a city scale in China. 

To the best of the authors' knowledge, there is no previous work to estimate the rooftop solar potential at a country scale for Lebanon. The presented work is thus the first attempt to produce  Lebanese urban and solar rooftop potential maps. We used a deep learning model to segment rooftops from high-resolution satellite imagery (50cm/pixel).

\hl{Unlike the work in \mbox{\cite{case_bei}\cite{solar_india}} which only relies on buildings' footprint area for solar potential estimation, we devise a greedy algorithm to simulate photovoltaic (PV) panels' placement based on each rooftop morphology. Our proposed panels placement algorithm processes regularized buildings' footprints and thus return more accurate results compared to \mbox{\cite{solar_ref1}}\mbox{\cite{solar_china}}.}

\begin{figure}[t]
\begin{center}
   \includegraphics[width=1.1\linewidth]{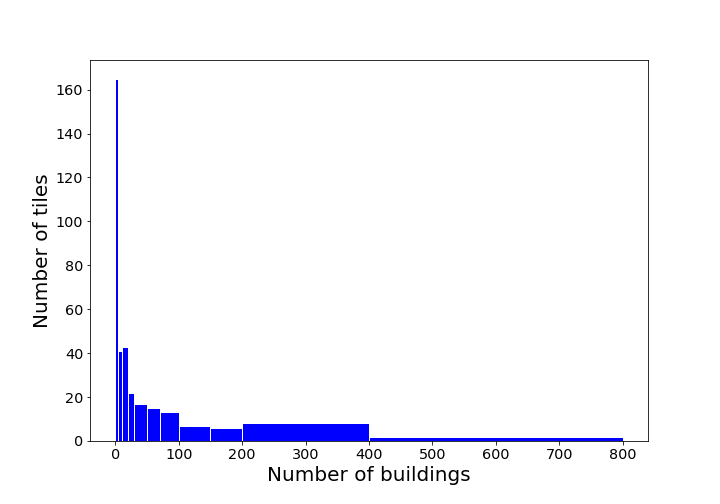}
\end{center}
   \caption{Tyre Urban Map (TUM) dataset tiles' distribution per buildings count.}
\label{Stats_dataset}
\end{figure}

\section{Dataset Creation}
\label{studyarea}
We aim in this research to build an efficient and accurate building segmentation model from Lebanese satellite images. Since neural networks are not guaranteed to generalize well on Out-Of-Distribution (OOD) data, a training dataset covering Lebanese populated areas with accurate buildings ground truth labels is needed. Such a dataset does not exist to our knowledge, so we created our own for this project.

\hl{The training dataset should hold a representative distribution of building types from all over Lebanon. Thus, we chose the Tyre area since it includes urban, suburban, rural, and slum areas. Tyre city itself is a dense urban city that lies within the outer ring of residential suburbs. Moreover, the chosen area has a refugee camp and vast countryside areas. Hence, the annotated training dataset does well represent the various characteristics of Lebanese areas. A segmentation model trained using this dataset would generalize well to the majority of Lebanese scenes.} 
 
We first chose a 35372x28874 GeoTIFF image with RGB channels taken by the GeoEye-1 satellite sensor of 50 cm/pixel resolution covering the Tyre district in the Southern Lebanon. We cropped the chosen area of interest into 1024x1024 chips, resulting in 338 tiles for manual annotation using the VGG-Image-Annotator tool \cite{VIA}. Finally, we further cropped each tile into non-overlapping 512x512 sized images while preserving each image's relative labeled polygons indices. \hl{We used non-overlapping tiles for the training dataset to avoid any redundant data which might cause model overfitting.} The final training dataset includes 1,352 tiles of 512x512 dimensions with ~10,000 buildings' objects. We refer to this dataset as Tyre Urban Map (TUM) dataset.

TUM has a balanced distribution of tiles with and without buildings (the ratio is roughly $(9:10)$). Figure \ref{Stats_dataset} portrays the number of tiles present in the TUM dataset for each given range of buildings' count. The average count is 28 buildings per tile. Moreover, Figure \ref{Stats_dataset} shows that the tiles are distributed over areas with varying building densities ranging from rural to suburban and urban regions. Thus, the created TUM dataset is expected to generalize well for the different aspects of building land cover distributions present in Lebanon.

As for the test set, we selected 30 areas of interest (AOI) from different Lebanese regions, including Beirut, Saida, Jounieh, Jbeil, and Tripoli. Selected AOI's encompass dense urban regions (Beirut), structured urban regions (Saida), and some rural areas. This diversity would help assess the generalizability of our model over the whole Lebanese territory. \hl{It is worthy to note that we addressed, at inference time, the issue of rooftops present at tiles' edges. Test images are cropped into 1024x1024 overlapping chips with a stride value of 512. Segmentation masks are averaged, resulting in smoothly merged buildings' footprints at the edges of the tiles.}

\begin{figure}[t]
\begin{center}
   \includegraphics[width=1.0\linewidth]{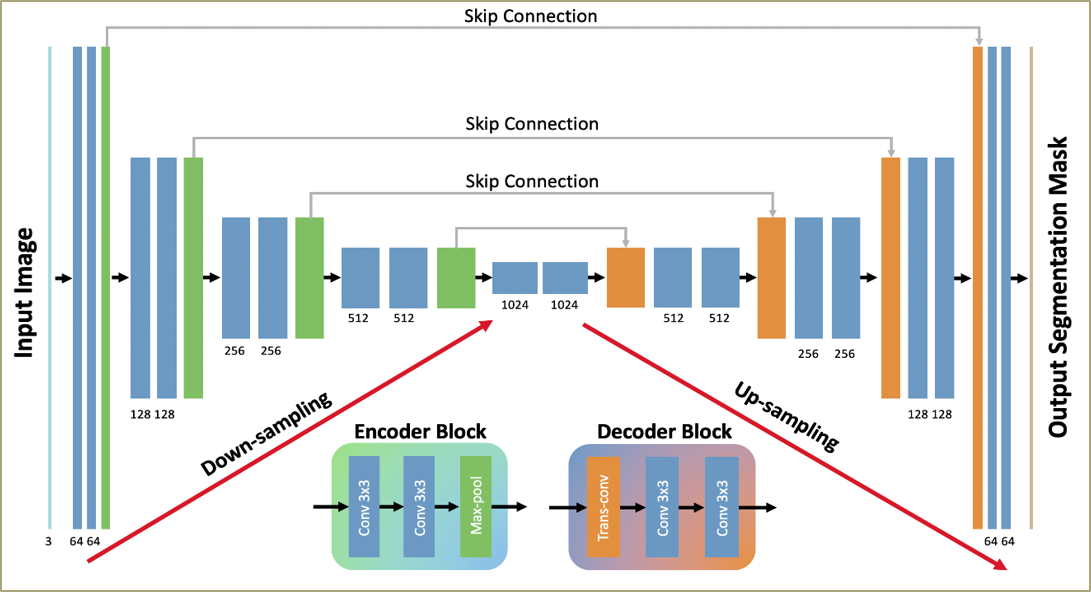}
\end{center}
   \caption{Encoder-Decoder UNet Architecture used with skip connections.}
\label{unet_fig}
\end{figure}

\section{Buildings' Footprints Map}
\label{method}

This section describes our approach to extracting buildings' rooftop polygons from satellite imagery of Lebanon. We adopt a UNet-like architecture to output semantic segmentation masks of the rooftops. However, semantic segmentation of rooftops is not enough to separate very close buildings. Hence, we also predict semantic segmentation masks for the buildings' boundaries. The building and boundaries masks are post-processed using the watershed algorithm \hl{\mbox{\cite{watershed_paper}}} to separate and tag each polygon with a unique identifier. Following this process, we obtain instance segmentation masks.

\subsection{Model Architecture}
UNet\cite{unet} is an end-to-end, fully convolutional neural network that serves the purpose of semantic segmentation of input images over multiple classes. It consists of a contracting path [Encoder] and an expansive path [Decoder] with skip connections between symmetrical blocks of identical size. The encoder part is comprised of repeated down-sampling convolutional and pooling layers followed by an activation function at every stage (i.e. RELU) to output feature maps of higher semantics and lower resolution. In contrast, the decoder is made up of consecutive up-sampling blocks. Skip connections allow information to flow directly from the low level to high-level feature maps \cite{ternausv2}. The overall architecture is a $U$-shaped encoder-decoder architecture as shown in Figure \ref{unet_fig}.

\begin{figure*}[ht!]
    \begin{center}
        \subfloat[Image]{\includegraphics[width = 0.25 \linewidth]{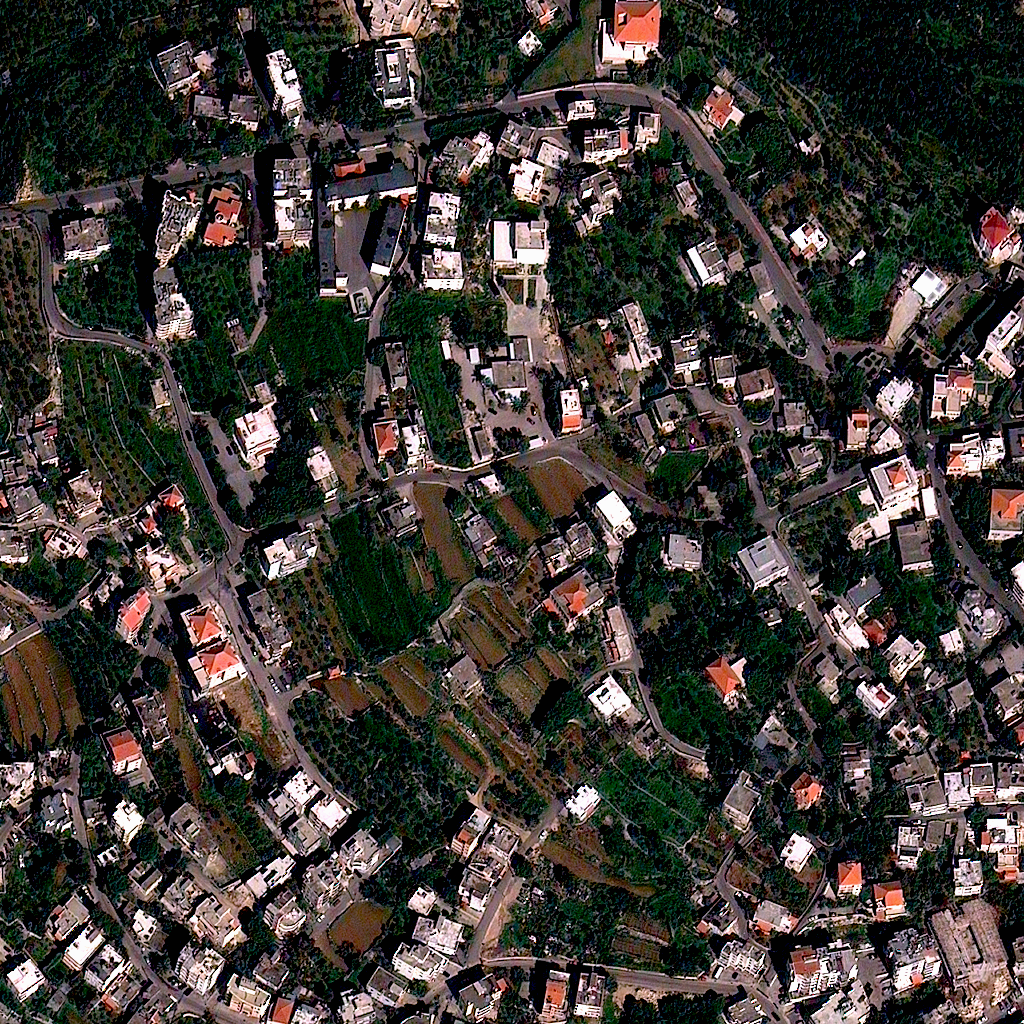}}
        \subfloat[Ground truth]{\includegraphics[width =  0.25 \linewidth]{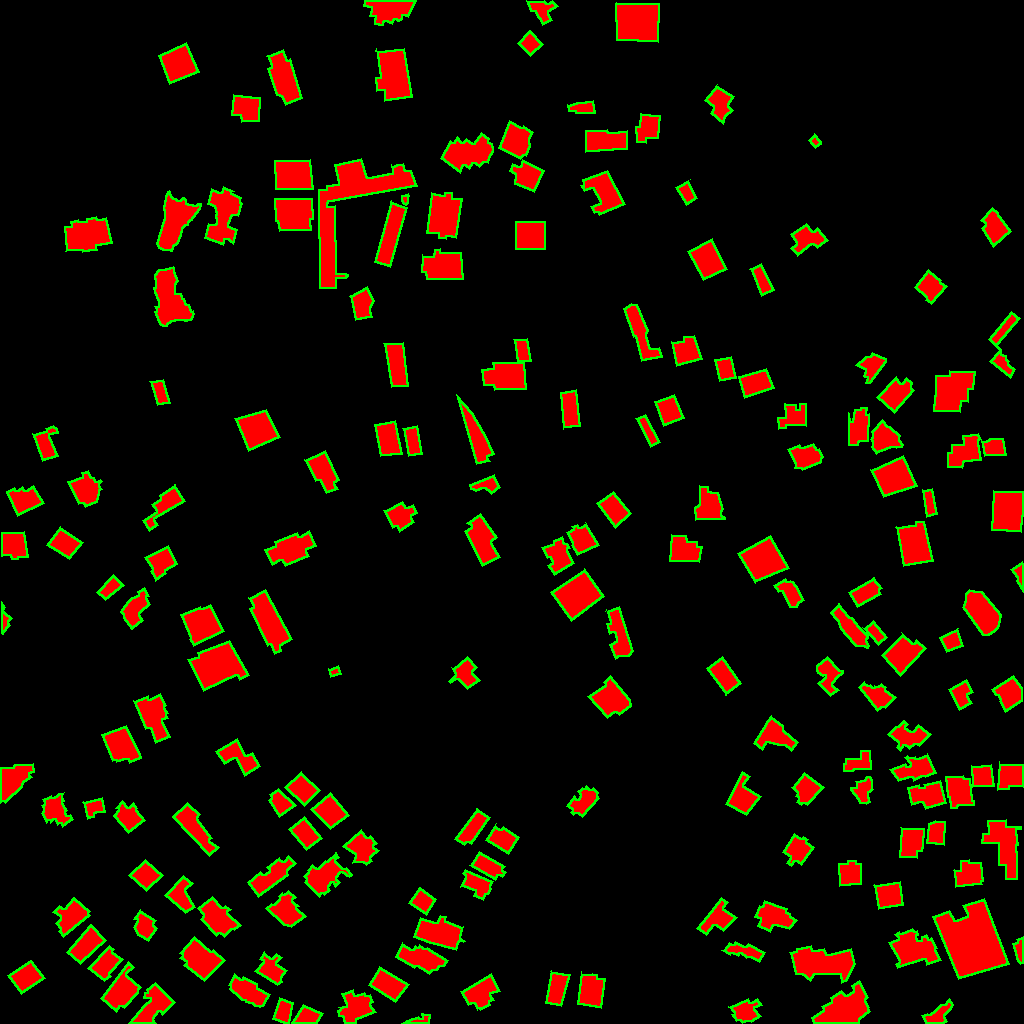}}
        \subfloat[Prediction mask]{\includegraphics[width =  0.25 \linewidth]{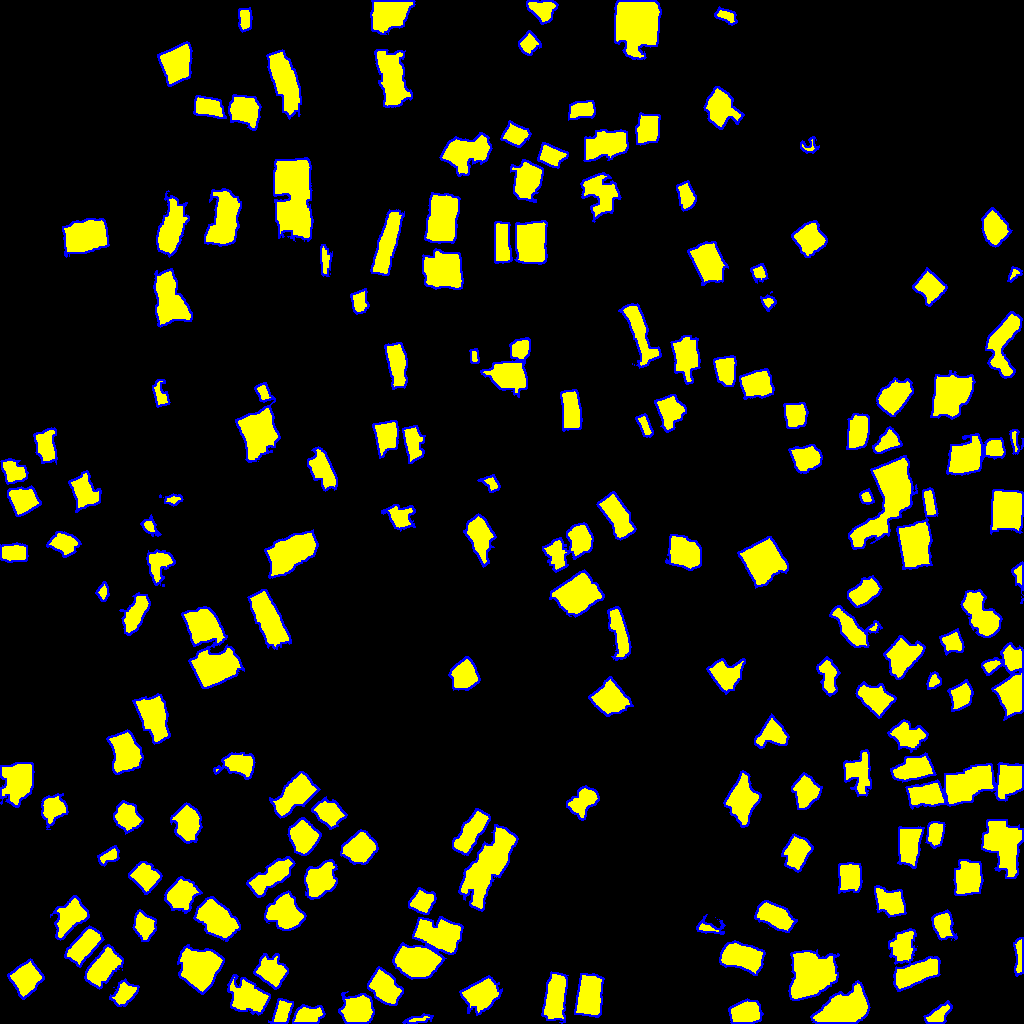}} 
        \subfloat[Prediction instance]{\includegraphics[width =  0.25 \linewidth]{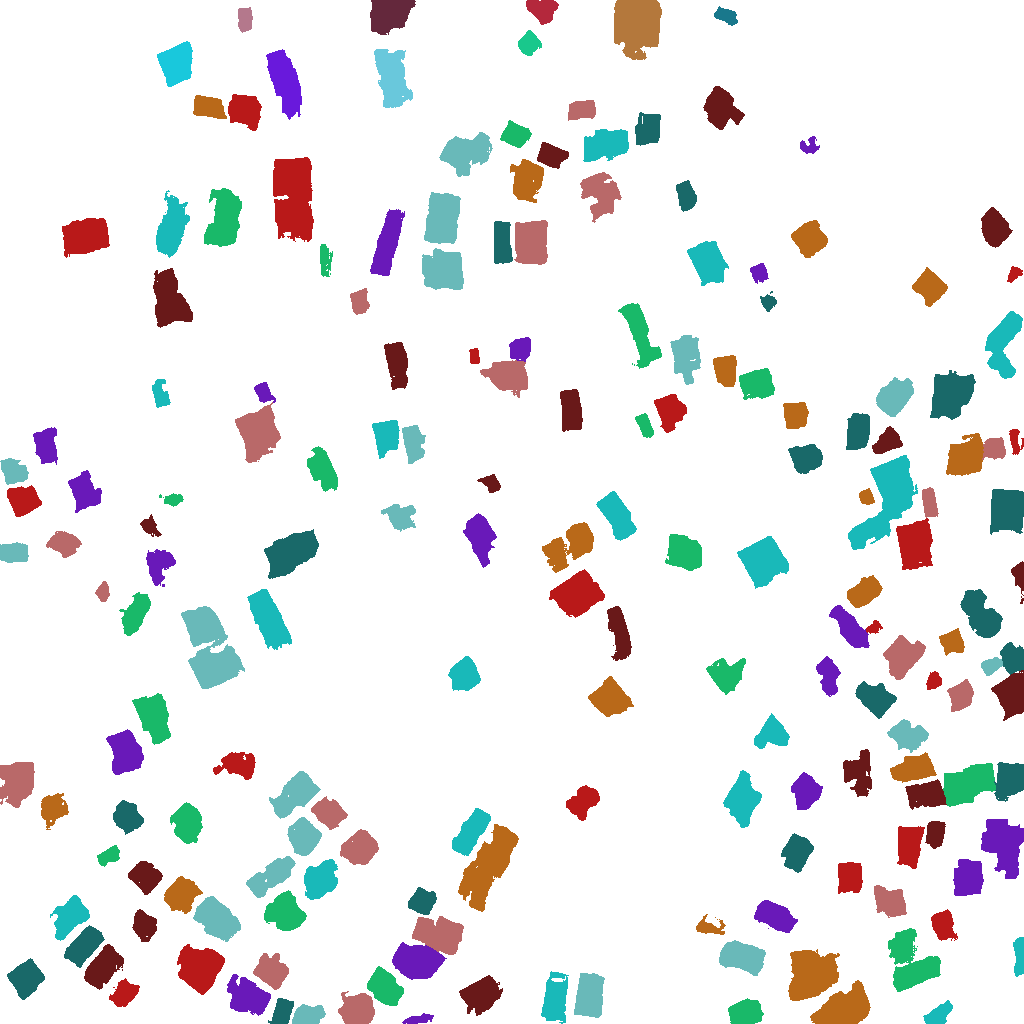}} \\
    \end{center}
    \caption{(a) Sample test image taken from Jounieh City, with (b) it's corresponding ground truth buildings mask (in red) and (c) our model predicted buildings' mask (in yellow) and finally (d) post-processing results for buildings instances segmentation .}
\label{segmentation_results}
\end{figure*}

The Encoder part of the architecture can be chosen from powerful and deep convolutional neural networks like Residual Nets \cite{resnet, resnext}, Inception Nets \cite{inception}, dual-path nets \cite{dpn}, or the newly introduced Efficient-Nets \cite{effnet}. In our experiments, Efficient-Net-B3 was found to perform better, in terms of accuracy and variance, than other members of the Efficient-Net family members and other encoders like ResNet34, ResNeXt50, InceptionV4, InceptionResNetV2 and DPN92 as shown in Table \ref{tbckbn_compare}, where variance is defined as the Fscore difference between the training and validation datasets.The model trained with EfficientNetB3 encoder had the best performance $(Fscore=84.3\%)$ and the best fitting on the train dataset as it achieved the lowest train/validation variance of $2.88\%$. The runner up model in terms of performance, InceptionV4, with $Fscore = 84.0\%$ had the highest (worst) variance of $4.95\%$. Generally speaking, all trained models achieved well in terms of $Fscore$, however EfficientNets showed better fit on the dataset with lower variance than other encoders.

\begin{table}[ht]
\begin{center}
\begin{tabular}{S|SS}\toprule
    {Backbone} & {F-score (\%)} & {Variance (\%)} \\\midrule
    {ResNet34} & {82.8} & {4.33} \\
    {ResNeXt50} & {83.7} & {4.24} \\
    {Inception-ResNetV2} & {84.0} & {3.55} \\
    {InceptionV4} & {84.1} & {4.95} \\
    {DPN92} & {83.8} & {3.74} \\
    {EfficientNetB2}  & {83} & {2.93} \\
    {\bf EfficientNetB3}  & {\bf 84.3} & {\bf 2.88} \\
    {EfficientNetB4}  & {83.8} & {2.97}\\ \bottomrule
\end{tabular}
\end{center}
\caption{F-score and Variance percentages on the validation set for different backbones \hl{trained for 100 epochs using identical hyperparameters}. It is evident that EfficientNetB3 achieves best performance in terms of F-score and variance.}
\label{tbckbn_compare}
\end{table}

As for the loss function, we chose a normalized weighted sum of losses across all output channels. Each single channel loss is a combination of Dice loss and Binary Cross-Entropy loss \cite{segloss} in order to leverage the benefits of both. \hl{It is worthy to mention that we investigated a combination of Dice loss and Focal loss using EfficientNetB3 and EfficientNetB4 backbones. However, no improvement was noticed where EfficientNetB3 F-score was equal to 84.2\% and EfficientNetB4 F-score equals 83.6\%.}

\subsection{Training Pipeline}
\label{tr}
\hl{All models were trained for 100 epochs} using Adam Optimizer \cite{adam} and the One-Cycle learning rate policy \cite{oclr} starting with an  initial learning rate = $\frac{0.0001}{20}$ and increases for 40 epochs in a cosine annealing manner till it reaches a maximum of 0.0001 and then decreases for the rest 60 epoch in the same annealing fashion. \hl{For each experiment, we save  weights' file of the best epoch with the highest F-score on the validation set.}

During training, we used mixed precision technique and applied random augmentations. We also used CutMix \cite{cutmix} data augmentation to further enhance our model robustness. We also used inference-time augmentations and employed post-processing to extract instance masks from semantic segmentation. The details of those different steps are beyond the scope of this manuscript.

\begin{figure}[ht!]
\begin{center}
   \includegraphics[width=1.0\linewidth]{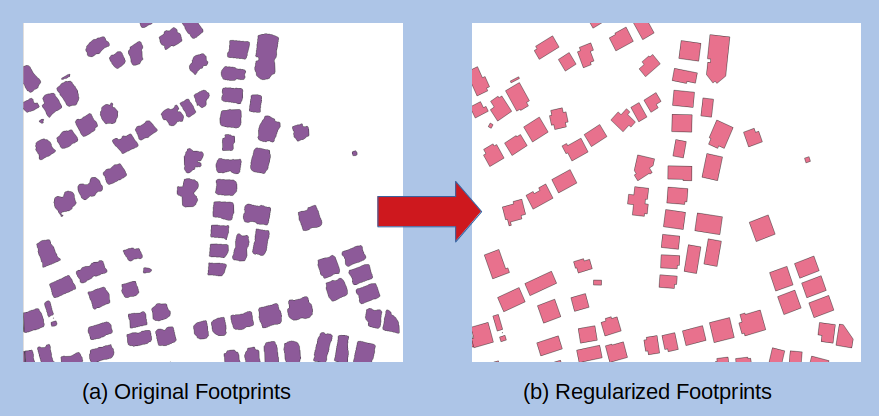}
\end{center}
   \caption{Buildings' footprint polygonization and regularization process: (a) model original output masks are first transformed into polygons and then (b) regularized to achieve buildings-like shape.}
\label{regularization_pic}
\end{figure}

Figure \ref{segmentation_results} shows the buildings' footprints extraction model in practice applied to Jonieh City in Lebanon, with it's corresponding ground truth building mask (in red) and predicted building mask (in yellow). Finally, we polygonized and regularized the output instance masks to achieve eye-pleasing regular-shape polygons as shown in Figure \ref{regularization_pic}.

\section{Solar Rooftop Potential Assessment}
\label{solar}

\begin{figure*}[hbt]
\begin{center}
   \includegraphics[width=0.7\linewidth]{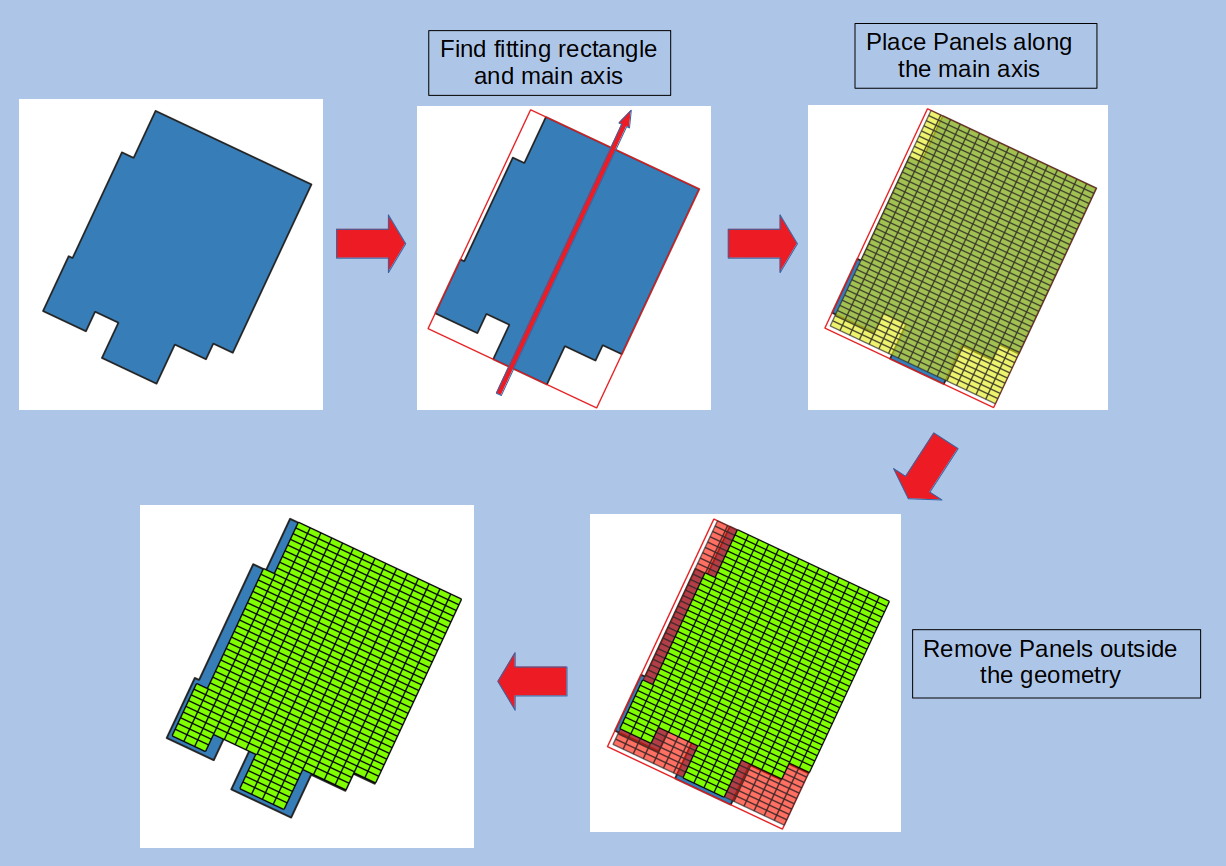}
\end{center}
   \caption{Solar panels fitting algorithm. The algorithm takes into consideration the morphology and the geometry of the roof. Solar panels are greedily placed along the longest axis of the building footprint to maximize the number of panels fitted inside the it's geometry. Panels extending outside the geometry are removed, and the remaining panels are taken into consideration for the solar potential calculation.}
\label{panel_placement}
\end{figure*}

The first step in calculating the solar potential of rooftops is to find how many solar panels can fit on each roof. However, due to the lack of critical information such as rooftops' relative slope and azimuth angle, we assume that all roofs are flat and at an orthogonal angle with the satellite sensor. Other factors like shading are also not considered due to insufficient information on the heights of all buildings in the country. In this work, we assumed a standard commercial solar panel module of dimensions (1x1.98) meters and a nominal power $P_{nominal}$ = 0.4 KWp. To estimate the number of panels that can fit on each rooftop, researchers such as \cite{case_bei} simply normalize the rooftop surface by the panel's area. This approach would result in an over-estimation of the solar rooftop capacity since it does not take the roof's morphology into account.

For the scope of this work, we devise the PV panels fitting algorithm described in Algorithm \ref{alg:fitting} based on the work presented in \cite{solar_ref1}. \hl{The PV panels fitting algorithm is applied to the regularized buildings' footprints to avoid any segmentation irregularities present at the roof edges.} Figure \ref{panel_placement} shows a running example of the panels fitting algorithm for a rooftop with extremely irregular shape.

\begin{algorithm}
\caption{Rooftop PV panels fitting algorithm.}
\label{alg:fitting}
\begin{algorithmic}[1]
\State For each building footprint polygon, find its minimum bounding rectangle $BBox$.
\State Designate longest axis of the $BBox$ as the main axis of the rooftop.
\State Place solar panels in a greedy way inside $BBox$ along the main axis.
\State Remove panels that extend outside the building polygon geometry. 
\end{algorithmic}
\end{algorithm}

\begin{figure}[ht]
\begin{center}
   \includegraphics[width=1.0\linewidth]{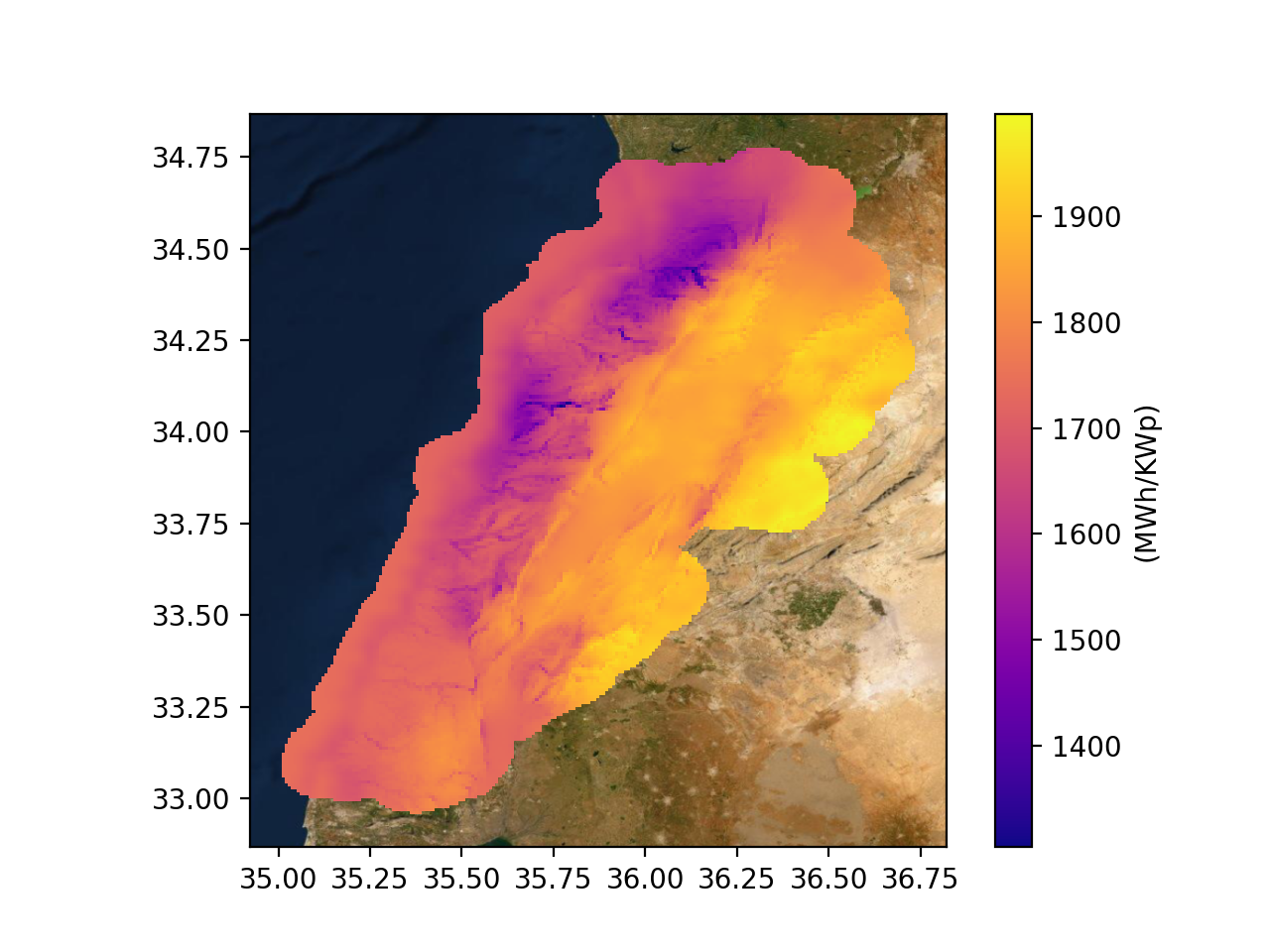}
\end{center}
   \caption{Lebanese $PV_{out}$ heatmap used for solar rooftop calculation. $PV_{out}$ is an average of the yearly production normalized to 1 $KWp$ of installed capacity. It is clear that areas within the East Beqaa Valley involve high potentials for solar energy generation.}
\label{pvout_heatmap}
\end{figure}

Finally, we use Equation \ref{eq:SolarPotential} to calculate, $SP_r$, the solar potential of every rooftop $r$ as follows:
\begin{equation}
SP_r (MWh/year) = N_r * P_{nominal} * PV_{out}
\label{eq:SolarPotential}
\end{equation}
where $N_{r}$ is the number of panels that fit on that roof, $P_{nominal}$ is the nominal power of the solar panel modules in ($KWp$), and $PV_{out}$ is the specific photovoltaic power output of the installation in ($MWh/KWp$).

$PV_{out}$ is defined as the amount of power produced per unit of the installed solar panel. In our studies, we used $PV_{out}$ data obtained from the World Bank Global Solar Atlas 2.0 \cite{Solargis}. The acquired data comprise a mapping of yearly averaged $PV_{out}$ values for every $~1Km^2$ grid tile across the whole Lebanese territory. The $PV_{out}$ heatmap for Lebanon is shown in Figure \ref{pvout_heatmap}. For every rooftop, we locate the corresponding tile within the $PV_{out}$ map and fetch its relative $PV_{out}$ value in $(MWh/KWp)$. Thus, we base our calculation on a more accurate and precise $PV_{out}$ map than other studies that use fixed $PV_{out}$ values \cite{case_bei}.

\begin{figure}[h]
\begin{center}
   \includegraphics[width=1.0\linewidth]{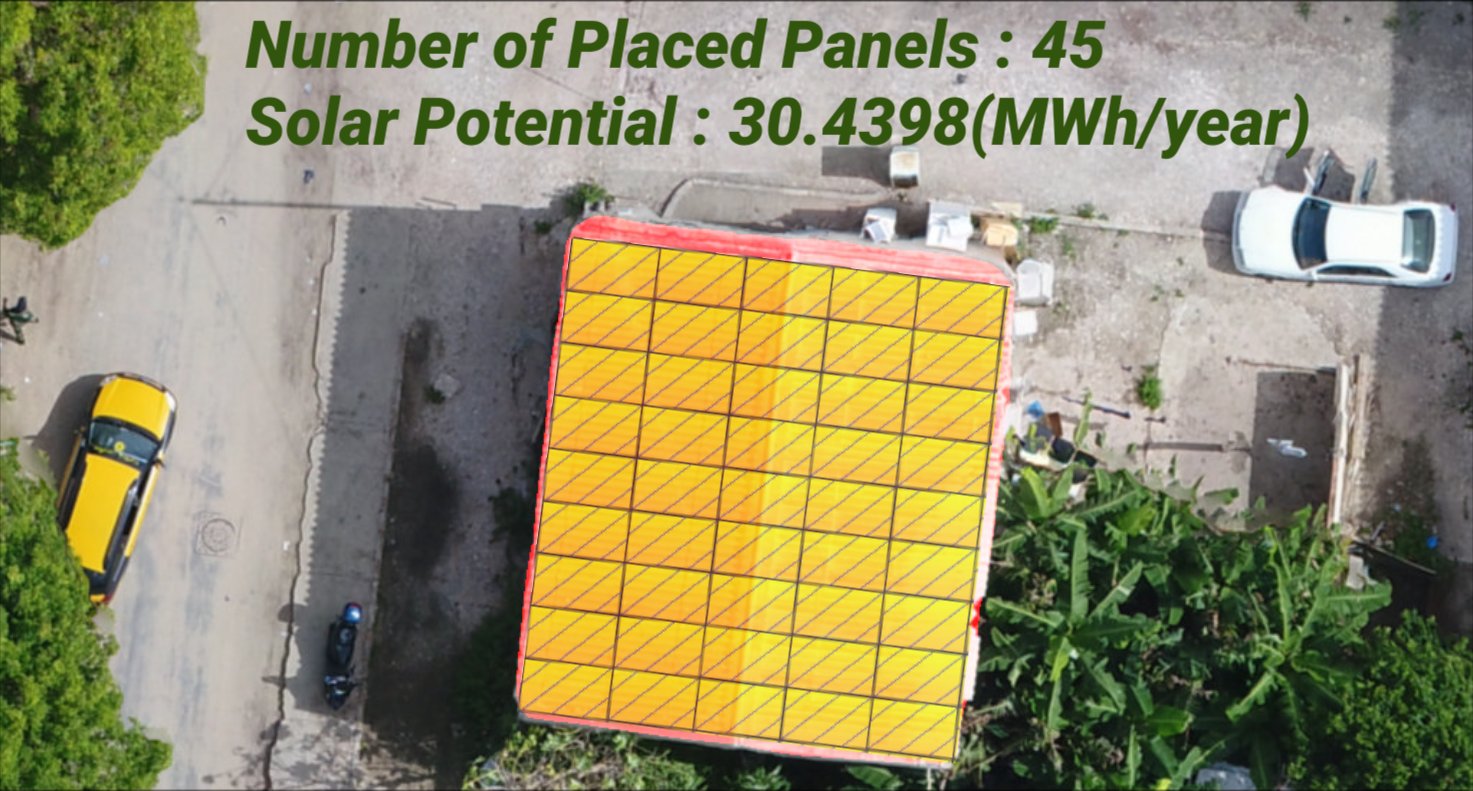}
\end{center}
   \caption{Panel fitting and solar potential calculation for a demo rooftop. The algorithm deduces that the roof maximum theoretical capacity holding  is 45 panels which corresponds to 30.4398 $MWh/year$. Small portion of the roof contours were left empty due to the inability of fitting a whole panel module in the remaining small area.}
\label{single_building}
\end{figure}

\section{Results}
\label{resuklts}
In this section, we present analysis and results for the solar rooftop potentials on two levels:  \textit{(i)} rooftop level and \textit{(ii)} district level. At the rooftop level, we reveal that the average roof's solar potential is sufficient enough to  cover most of the energy needs of Lebanese households and achieve a sense of energy security. At the district level, we compute the total and average solar rooftop potentials per district and compare these values with the maximum hypothetical capacity of each district. 

\subsection{Household Energy Security}
For every rooftop, we fetch its relative $PV_{out}$ value and fit the maximum number of panels. Since we use a greedy algorithm for the solar panels' placement, we then calculate the maximum theoretical solar rooftop potential capacity. Figure \ref{single_building} shows the PV panels fitting algorithm and solar potential calculation scheme in action for a demo house where 45 panels were placed which corresponds to 30.4398 $MWh/year$.

In practical situations, fewer solar panels could be placed due to \hl{various parameters, including objects usually available on the roof (such as roof tanks and solar water heaters, among others), inclined roof slope, and the requirement of leaving sufficient room for service and emergency access.} We  multiply the number of fitted solar panels with a Utilization factor $U$ to account for this effect. Although studies in the literature  \hl{\mbox{\cite{2013PHOTOVOLTAICLEBANON}}}\hl{\mbox{\cite{case_bei}}} report results for only 50\% and 75\%  utilization factors, for the scope of this work, we experimented with several utilization factors in the following results sub-sections, where $U \in [0.1, 0.25, 0.5, 0.75, 1]$.

\begin{figure}[h]
\begin{center}
\includegraphics[width=1.0\linewidth]{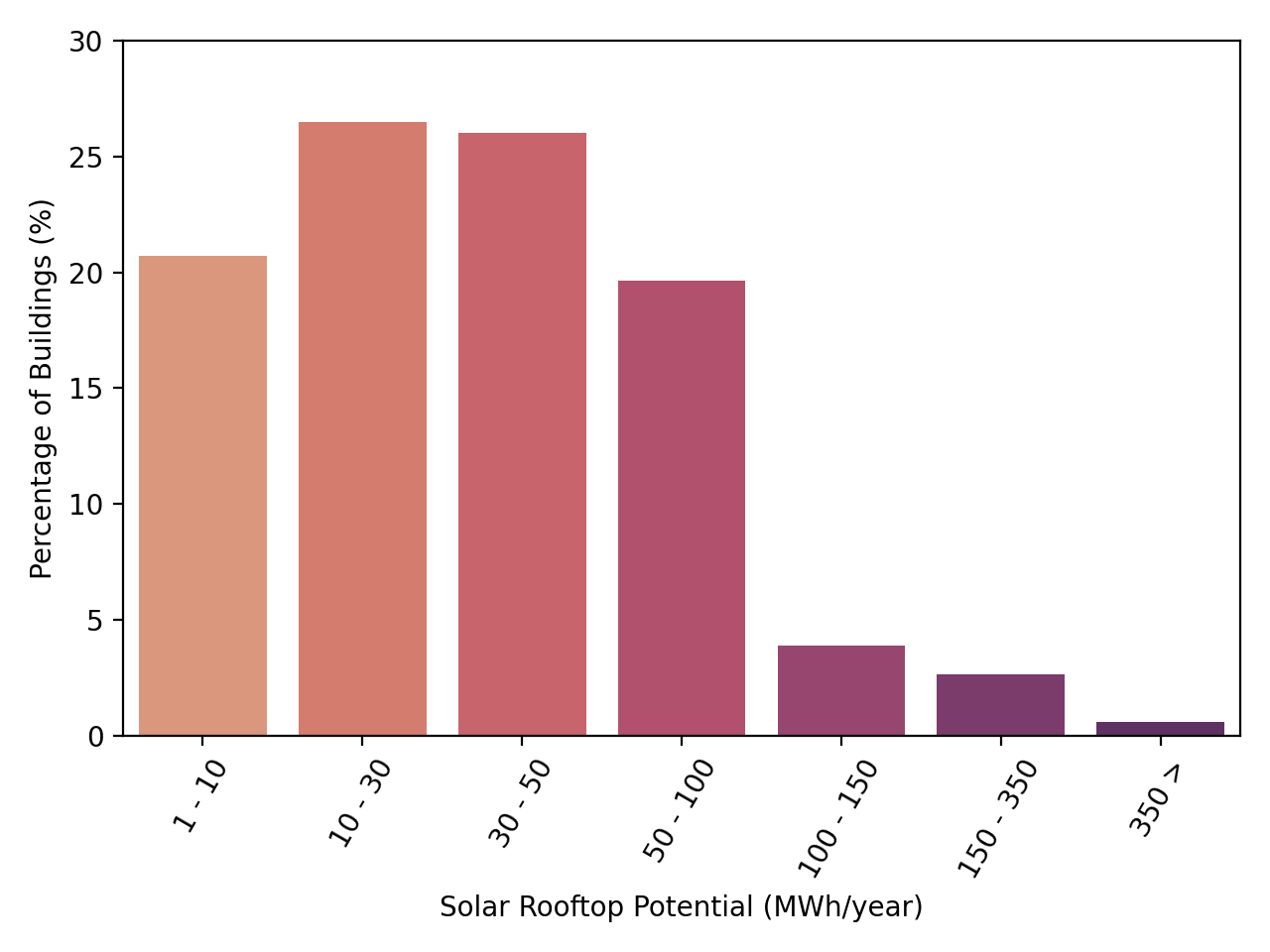}
\end{center}
\caption{Buildings' distribution across different bins of solar rooftop potential using $U = 50\%$. More than half of the rooftops can produce energy in the range \(\left[10, 50 \right[\) $MWh/year$. While remaining 40\% of the rooftops are equally distributed among the two intervals   : \(\left[1, 10 \right[\) and \(\left[50, 100 \right[\) $MWh/year$.}
\label{sp_count}
\end{figure}

The histogram presented in Figure \ref{sp_count} shows buildings' distribution across solar rooftop potential bins in $MWh/year$. Those results are computed assuming a 50\% Utilization factor, $U$.

Rooftops' solar potential that falls in the range \(\left[1, 10 \right[\) $MWh/year$ account for around 21\% of the distribution. Those are usually homes characterized with rooftops of a smaller surface or lower $PV_{out}$ value than the rest of the distribution.
On the other edge, 7\% of the buildings are associated with high expected solar rooftop potential, between 100 $MWh/year$ and 350 $MWh/year$. Values above 350 $MWh/year$ are primarily due to segmentation errors in high density areas such as slums where multiple nearby building footprints are merged.

\begin{table}[ht]
\begin{center}
\begin{tabular}{p{0.25\linewidth} | p{0.3\linewidth}  p{0.2\linewidth}}\toprule
    {Utilization Factor} & {Average rooftop Solar Potential} & {Standard Error} \\ \midrule
    {10\%} & {9.747} & {$\SI{\pm 6.22}{}$} \\ 
    {25\%} & {22.21} & {$\SI{\pm 14.87}{}$} \\
    {50\%} & {40.81} & {$\SI{\pm 27.67}{}$} \\
    {75\%} & {58.11} & {$\SI{\pm 38.54}{}$} \\
    {100\%} & {74.29} & {$\SI{\pm 48.09}{}$} \\\bottomrule
\end{tabular}
\end{center}
\caption{Average rooftop Solar Potential $(ASP)$ in $MWh/year$ per building in Lebanon and it's corresponding Standard Error in $MWh/year$ for various utilization factors. \hl{$ASP$ is defined as the mean of the solar rooftop potential of all buildings across Lebanon.}}
\label{avg_sp_std_leb}
\end{table}

More than half of the distribution (53\%  of the buildings) can leverage a total solar rooftop potential in the range \(\left[10, 50 \right[\) $MWh/year$. Finally, 19\% of the buildings have rooftops capable of producing a high total solar potential between \(\left[50, 100 \right[\) $MWh/year$.

\hl{For the scope of this work, Average Solar rooftop Potential $(ASP)$ is defined as the mean of the solar rooftop potential of all buildings across Lebanon.} Table \ref{avg_sp_std_leb} presents $ASP$ in $MWh/year$ for various utilization factors with the associated Standard Error $(SE)$ value. $SE$ is defined in Equation \ref{eq:SE}. Large observed $SE$ values are attributed to the wide variety of  solar rooftop potential in the buildings' distribution set as shown earlier in Figure \ref{sp_count}.

\begin{equation}
SE = \frac{\sigma}{\sqrt{n}}
\label{eq:SE}
\end{equation}

where $\sigma$ is the sample standard deviation and $n$ is the number of samples.

A recent survey \cite{AppliancesLb} shows that households in Lebanon consume on average 5.41 $MWh/year$. A low 10\% Utilization factor can accommodate an average of 9.747 $MWh/year$ which corresponds to 180\% of the average household's electricity consumption per year. \hl{Using only around 5\% of the roof surface, would allow single-family residences} to significantly cut their electricity bills and meet near-net zero energy building (near-$NZEB$) targets \cite{WEI2021100018}. And also, sell surplus energy back to the grid and generate additional income, once corresponding energy regulations are updated to allow for net-metering.

Economic crisis and severe inflation witnessed since 2019, made energy security a real concern for Lebanese households. Energy security is defined as the uninterrupted availability of energy sources at an affordable price. In Lebanon, both conditions are not currently met, where most families suffer from long hours of electricity outage daily, while the average electricity bill largely exceeds the minimum wage rate. Assuming a 50\% utilization factor scenario, the expected Average Solar rooftop Potential (ASP) of a building is 40.81 $(MWh/year)$, which is equivalent to the average yearly consumption of 7.5 households. Hence, residential apartments can achieve a significant percentage of energy security by relying on rooftop solar energy production.

\subsection{Urban Solar Potential Factors}
Lebanese territory is divided into 26 districts, where each district exhibits its own urban and topographic characteristics. As shown in Figure \ref{pvout_heatmap}, the specific photovoltaic power output of the installation, $PV_{out}$ values, are not uniformly distributed across districts. A wide $PV_{out}$ gap of more than 500 $MWh/KWp$ is observed between locations receiving the lowest and highest amount of solar irradiation. Hence, rooftops, with the same panel capacity, in different districts,  do not produce the same solar rooftop potential. 

\begin{figure}[!h]
\begin{center}
\subfloat[\label{Ttlsp_maps}]
{\includegraphics[width = 3in]{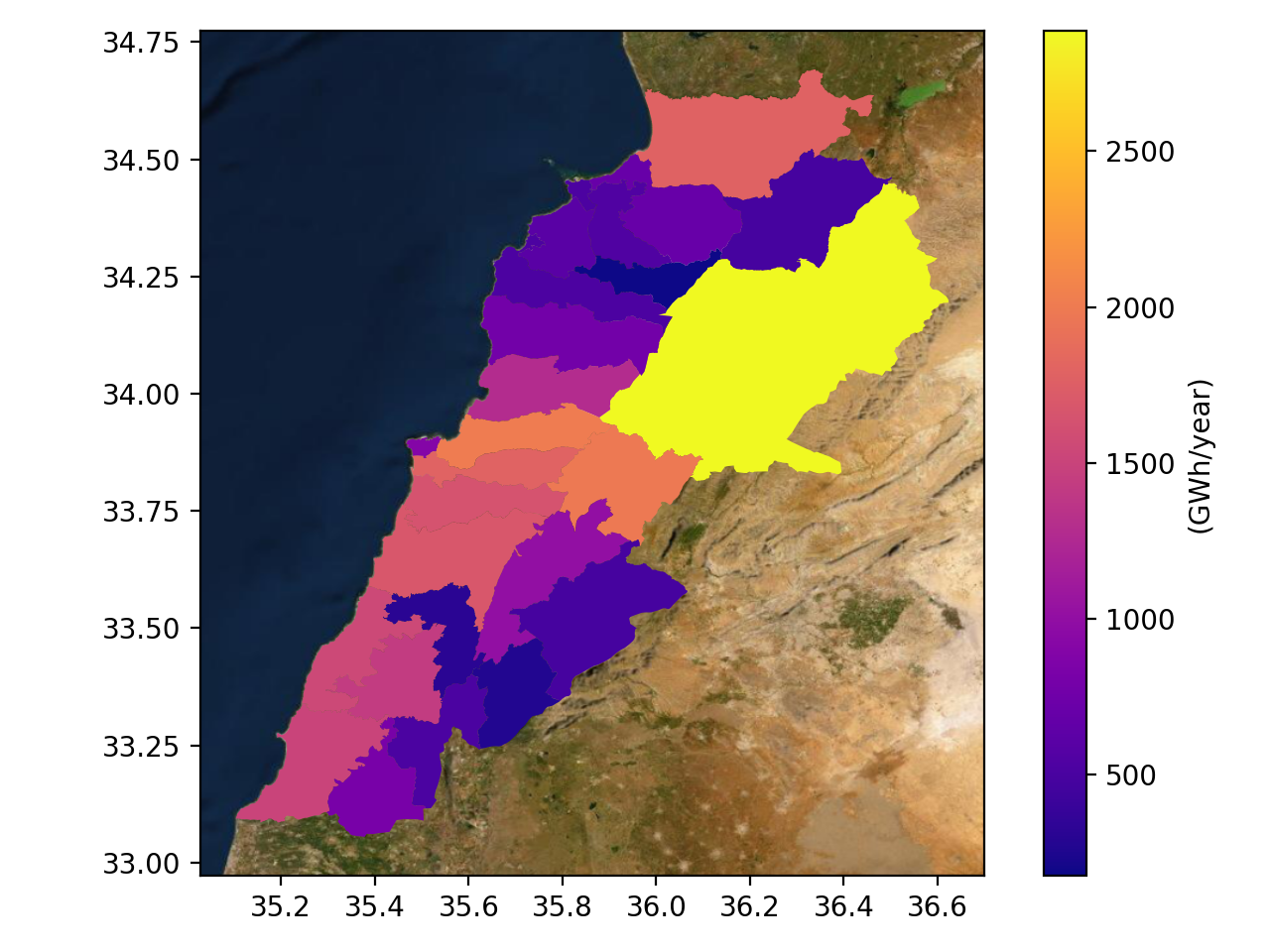}}\\
\subfloat[\label{hmap}]
{\includegraphics[width=3in]{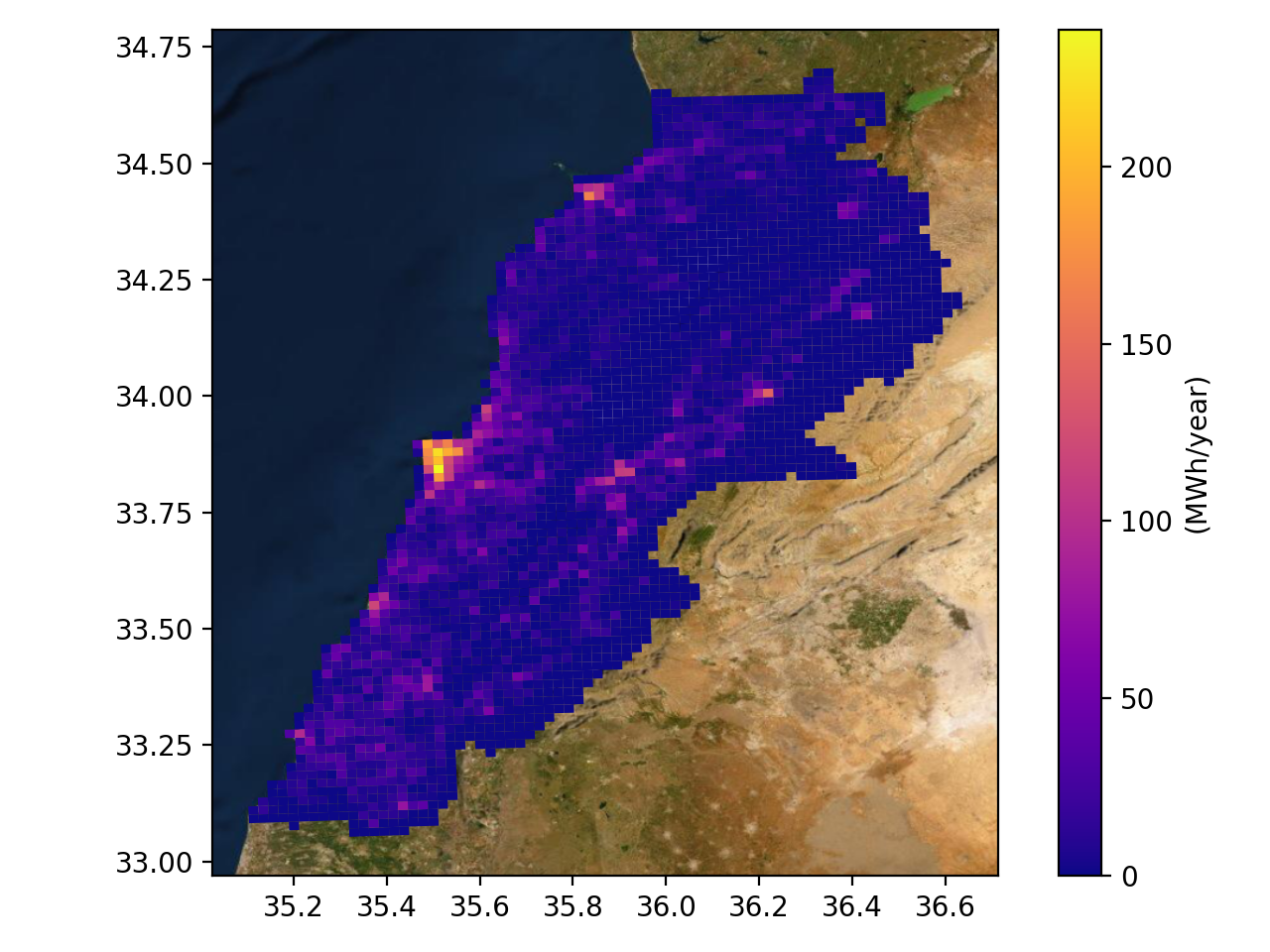}}
\end{center} 

\caption{(a) Total solar rooftop potential in $GWh/year$ for different Lebanese districts using 50\% Utilization factor. (b) Heatmap showing average solar rooftop potential $MWh/year$ using a sliding window of 4 $Km^2$ area.}
\label{sp_maps} 
\end{figure}

\begin{figure}[ht]
\begin{center}
   \includegraphics[width=1.0\linewidth]{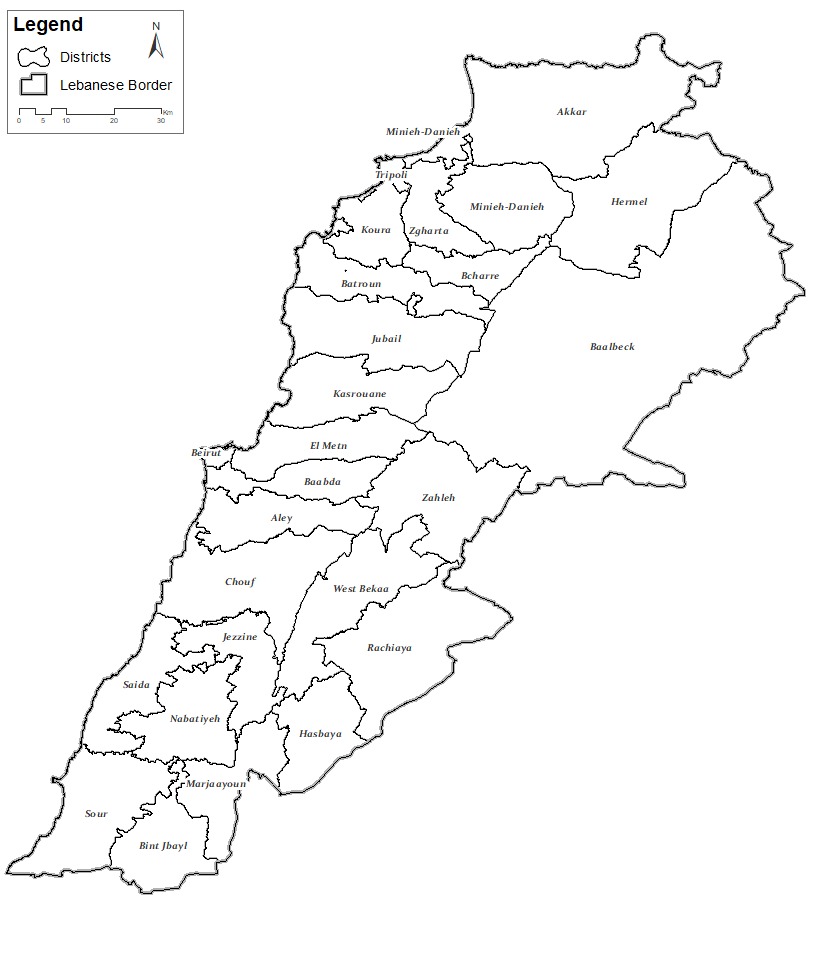}
\end{center}
   \caption{Lebanese districts' map [33.8547\degree N, 35.8623\degree E].}
\label{districts}
\end{figure}

We calculate total solar rooftop potential in $GWh/year$ for each district using a 50\% Utilization factor. We also calculate the average solar rooftop potential in $MWh/year$ using a sliding window of size (surface) $4Km^2$. 
We visualize those results in Figures \ref{Ttlsp_maps} and \ref{hmap}, respectively. Figure \ref{districts} is included as a reference figure to show the layout of the Lebanese districts.

Figure \ref{Ttlsp_maps} shows that Baalbeck (2,885 $GWh/year$), El-Metn (2,014 $GWh/year$) and Zahle (1,979 $GWh/year$) districts have the highest total solar rooftop potentials. On the other hand, in Figure \ref{hmap}, we observe that Baalbeck district is almost uniformly blue, which indicates that it is not well populated. Baalbeck, the largest district with the highest $PV_{out}$ values, is indeed expected to provide the highest total solar rooftop potential. However, El-Metn district's total solar rooftop potential is only 871 $GWh/year$ behind Baalbeck despite being 11 times smaller and witnessing lower $PV_{out}$ values. Baalbeck belongs to the most yellowish region in Figure \ref{Ttlsp_maps} while being the most bluish one in Figure \ref{hmap}.

\begin{figure*}[h]
\begin{center}
   \includegraphics[width=0.7\linewidth]{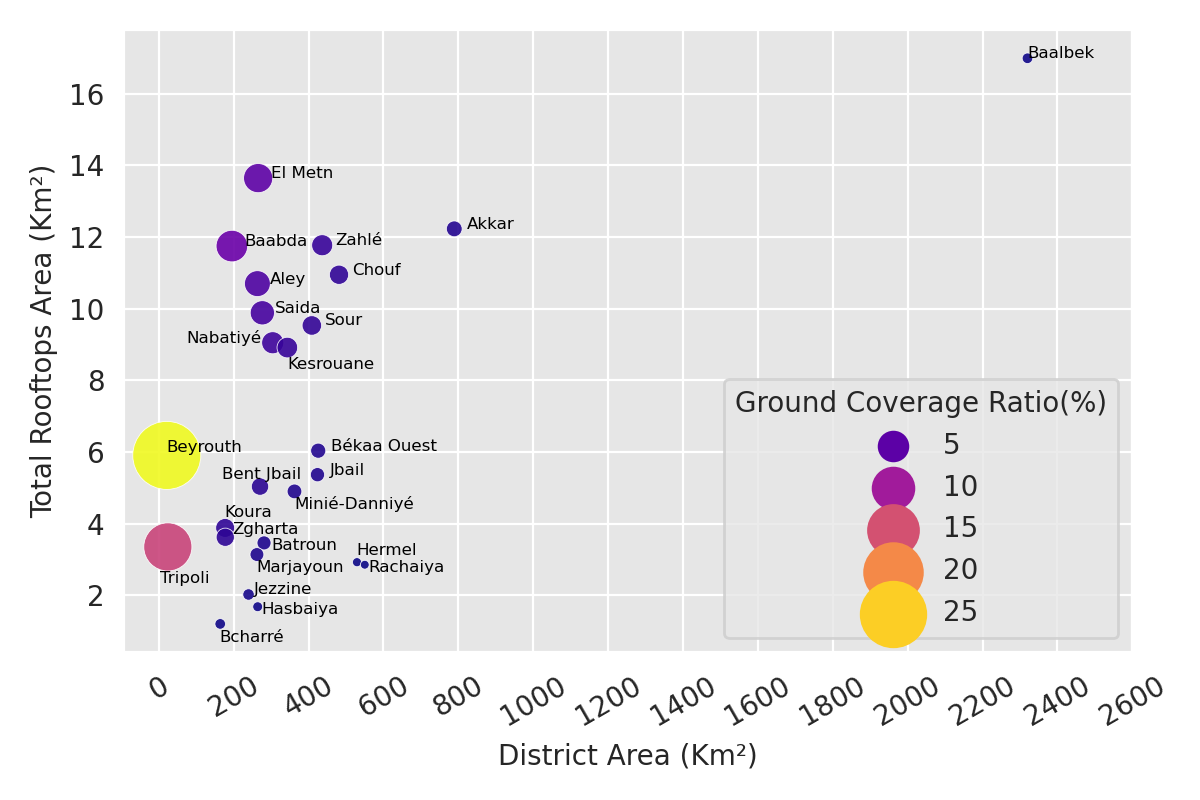}
\end{center}
   \caption{Bubble Chart showing the Ground Coverage Ratio ($GCR$)\cite{Ng2011ImprovingKong} for each district. The x- and y- axes show each district surface and its corresponding rooftops area in $Km^2$, respectively. The bubble size designates the $GCR$ value for each district. Beirut district shows the highest $GCR$ value of $27\%$, whereas the lowest $GCR$ value corresponds to Rachaiya district with only $0.55\%$.}
\label{bua_bubble_plot}
\end{figure*}

To further analyse those results, we plot in Figure \ref{bua_bubble_plot} a bubble chart that shows the Ground Coverage Ratio ($GCR$)\cite{Ng2011ImprovingKong} for each district. Ground Coverage Ratio, defined in Equation \ref{eq:gcr}, is used to assess the percentage of built-up area in each district:

\begin{equation} 
GCR = \frac{A_b}{A_T}= \frac{A_{Building Footprints}}{A_{District}}
\label{eq:gcr}
\end{equation}

Where $A_b$ is the built-up area or the total area of building footprints in a district, and $A_T$ is the domain area and for the context of this study, it is defined as the district area.\\

Baalbeck district has one of the lowest $GCR$ values of 0.73\%. EL Metn district on the other hand has 5.4\% $GCR$ value, which is $7x$ larger than Baalbek. This huge difference in built-up area justifies results observed in Figures \ref{Ttlsp_maps} and \ref{hmap}.

For the sake of clarity and to further expand this line of analysis, we define the maximum Hypothetical solar potential Capacity ($HC_i$) of each district $i$, as the maximum solar potential achievable at district $i$, if we lay solar panels across the district whole area. $HC_i$ is a hypothetical value, where we assume that the district topology is flat. $HC_i$  is used in the scope of this work as an indicator to assess the amount of solar potential that can be produced at each district level. $HC_i$ values are reported in Table \ref{tb6}.

Baalbeck, Lebanon's largest district, has a maximum hypothetical capacity of 845 $TWh/year$. However, due to the low built-up area in the region, solar potential cannot be efficiently exploited at the rooftop level. An alternative solution we suggest here is to install large solar farms in Baalabek district to avail solar potentials.  

\subsection{District-Level Analysis}
\hl{Beirut solar map \mbox{\cite{case_bei}} reports 195 MWp as the nominal capacity of the city for a 50\% Utilization factor and 295 MWp for a 75\% Utilization Factor. Our findings shows that the nominal capacity of the city is 285 and 427 MWp for 50\% and 75\% Utilization factors, respectively.
The difference is due to the following factors: \textit{(i)} assumption in \mbox{\cite{case_bei}} that every 8 sqm yields a total of 1 KWp. While in our work, we used a standard commercial PV panel with an area equal to 1.98 sqm and nominal power equal to 0.4 KWp, which is 1.6x larger. Thus, results in \mbox{\cite{case_bei}} should be weighted by a 1.6X factor for a fair comparison. 
\textit{(ii)} The difference between the newly weighted values (312 and 472 MWp) and our findings (285 and 427 MWp) is less than 10\%. \textit{(iii)} PV panel placement is not employed in \mbox{\cite{case_bei}} and thus roof morphology is not taken into consideration which ended up with an overestimation of the actual solar rooftop potential of the city.}

The capital city (Beirut) and the economic capital (Tripoli) have fairly mid-range total rooftop solar potential with 907 and 516 $GWh/year$, respectively. These two districts have the most glowing spots on the heatmap shown in Figure \ref{hmap} and exhibit the highest $GCR$  percentages ($27.7\%$ and $13.9\%$, respectively) as shown in Figure \ref{bua_bubble_plot}. 

Being extremely populated areas, Beirut and Tripoli districts have the highest average solar rooftop potential using a sliding window of surface $4Km^2$ as shown in Figure \ref{hmap}. However, if shadowing effects were considered, one would expect lower average solar rooftop potential in these two high-density districts. 

Baabda (1,786 $GWh/year$), Aley (1,641 $GWh/year$), Saida (1,547 $GWh/year$), Tyr (1,503 $GWh/year$) and Nabatiye (1,430 $GWh/year$) districts show a large distribution of pink spots in Figure \ref{hmap}. These districts are mid-populated areas ($GCR$ values between 4\% and 6\%) with high $PV_{out}$ values between 1,700 and 1,800 $MWh/KWp$. Using a 50\% utilization factor, the total solar rooftop potential is fairly high in these districts and reach 3\% to 4\% of their maximum hypothetical capacity as shown in Table \ref{tb6}.

Finally, Hermel, Hasbaya, Jezzine, Rachaiya and Bcharre districts experience a blue tone in both Figures \ref{Ttlsp_maps} and \ref{hmap}. Average solar rooftop potential in those districts is poorly utilized, with less than 0.7\% $GCR$ value. Total solar rooftop potential does not exceed 1\% of each district's maximum hypothetical capacity.

We also provide in Table \ref{tb6} detailed average and total solar rooftop potential for every district while varying the utilization factor from 10\% to 100\%. \hl{Assuming a 50\% utilization factor, the total solar rooftop potential for all districts sums up to 28.1 $TWh/year$. Whereas Lebanon total energy consumption for the year 2019 (including Electricity of Lebanon, EDL, and private generators) was estimated by our EDL expert contact to be around 12.5 $TWh/year$. In the last column, we report $\%\frac{TSP}{HC}$ which constitutes the relative percentage of Total solar rooftop potential (at $U$ = 100\% ) with respect to the maximum hypothetical capacity for each district. It is worthy to note that $\%\frac{TSP}{HC}$ is directly proportional to $GCR$.}

\section{Conclusion}
\label{conclusion}
We discussed in this manuscript buildings' footprints segmentation for Lebanon from satellite imagery. We also computed solar rooftop potential using a panel fitting algorithm to produce the first complete solar potential map of Lebanon. Furthermore, we conducted rooftop- and district- level analysis to deduce patterns and provide policymakers and key stakeholders with tangible insights to design tailored regulations and future directions. \hl{We showed that 5\% of the roof surface would be enough to accommodate the yearly electric needs of a single-family residence. As for residential apartments, the average solar rooftop potential using a 50\% utilization factor is sufficient to provide up to 8 households with energy security. Baalbek district holds the highest total solar rooftop potential, while Beirut reports the highest average solar rooftop potential. We found that Lebanon's total solar rooftop potential is 28.1 $TWh/year$ assuming a 50\% utilization factor, which is more than double the national energy consumption for 2019. Finally, we found that low-populated districts failed to deliver more than 1\% of their maximum hypothetical capacity. Large solar farms are highly recommended solutions for Baalbek and Hermel districts to avail the high $PV_{out}$ potentials.}

Although our work leverages more accurate estimates of the rooftop potential than previous works for Lebanon, certain improvements can be further conducted concerning the buildings footprint extraction, panel placement algorithm, \hl{assessment of the actual roof utilization factor} and the $PV_{out}$ map. For instance, in addition to the segmentation masks we can estimate the building heights and the orientation of the points on the rooftops surface. This would eliminate the zero-shadowing and the orthogonal flat rooftop assumptions. Based on the heights, we can take shadowing effect between buildings into consideration. We can also reconstruct a 3D surface of the rooftops given the orientation parameter, which would yield in a better estimate of the total available area of the rooftop, it's morphology and the presence of any obstacles. The panel placement algorithm can be improved by aligning the panels along the direction that maximizes the energy production as done in practice instead of aligning them along the main axis of the rooftop. Rooftop obstacles, best PV panel installation angle, and the spacing between the panels can be also taken into consideration to find the best distribution of the panels for each rooftop. \hl{An assessment study to find the actual rooftop utilization factor in Lebanon is essential to report practical figures. Such assessment would require first to generate ground truth labels using either remote sensing or field visits.} Finally, the extraction of a higher resolution $PV_{out}$ map is also key to attain more accurate results.

{\small
\bibliographystyle{ieee_fullname}
\bibliography{references}
}

\clearpage
\onecolumn

\begin{center}
\begin{longtable}{c||ccccc|ccccc|c||c} \toprule
   \multirow{2}{*}{}  & \multicolumn{5}{c}{\multirow{2}{*}{\thead{Average Solar rooftop\\ Potential ($ASP$ in $MWh/year$)}}}&\multicolumn{5}{c}{\multirow{ 2}{*}{\thead{Total Solar rooftop\\ Potential ($TSP$ in $GWh/year$)}}}&\multicolumn{1}{c}{\multirow{ 2}{*}{\thead{Hypothetical Capacity\\ ($HC$ in $TWh/year$)}}}&\multirow{2}{*}{\thead{$\%\frac{TSP}{HC}$}} \\\\
   \cmidrule{1-13}

   \multirow{2}{*}{District}&\multicolumn{10}{c|}{Utilization Factor}&\multicolumn{1}{c}{\multirow{2}{*}{}}&\multirow{2}{*}{} \\ %
   {}&{$10\%$}&{$25\%$}&{$50\%$}&{$75\%$}&{$100\%$}&{$10\%$}&{$25\%$}&{$50\%$}&{$75\%$}&{$100\%$}&\multicolumn{1}{c}{}&{} \\ \cmidrule{1-13}
   {Baalbek}&{7}&{17}&{35}&{53}&{70}&{577}&{1442}&{2885}&{4328}&{\bf 5771}&{\bf 845}&{0.68} \\
   {El Metn}&{9}&{22}&{45}&{68}&{91}&{402}&{1007}&{2014}&{3022}&{4029}&{74}&{5.44} \\
   {Zahlé}&{9}&{24}&{49}&{74}&{98}&{395}&{989}&{1979}&{2968}&{3958}&{139}&{2.85} \\
   {Baabda}&{10}&{27}&{54}&{81}&{109}&{357}&{893}&{1786}&{2679}&{3572}&{50}&{7.14} \\
   {Akkar}&{6}&{15}&{31}&{46}&{62}&{355}&{887}&{1775}&{2662}&{3550}&{235}&{1.51} \\
   {Chouf}&{7}&{18}&{37}&{56}&{75}&{335}&{839}&{1679}&{2519}&{3359}&{141}&{2.38} \\
   {Aley}&{9}&{24}&{48}&{73}&{97}&{328}&{820}&{1641}&{2461}&{3282}&{74}&{4.44} \\
   {Saida}&{8}&{20}&{40}&{60}&{80}&{309}&{773}&{1547}&{2321}&{3095}&{73}&{4.24} \\
   {Tyr}&{7}&{18}&{36}&{54}&{73}&{300}&{751}&{1503}&{2255}&{3006}&{124}&{2.42} \\
   {Nabatiyé}&{6}&{16}&{33}&{50}&{67}&{286}&{715}&{1430}&{2145}&{2860}&{87}&{3.29} \\
   {Kesrouane}&{7}&{17}&{35}&{52}&{70}&{255}&{639}&{1278}&{1917}&{2556}&{98}&{2.61} \\
   {Békaa Ouest}&{9}&{23}&{47}&{71}&{95}&{201}&{504}&{1009}&{1514}&{2019}&{126}&{1.60} \\
   {Beyrouth}&{18}&{46}&{93}&{139}&{\bf 185}&{181}&{453}&{907}&{1360}&{1814}&{\bf 4}&{\bf 45.35} \\
   {Bent Jbail}&{6}&{17}&{34}&{51}&{68}&{163}&{409}&{818}&{1227}&{1637}&{80}&{2.05} \\
   {Jbail}&{5}&{14}&{28}&{42}&{56}&{155}&{389}&{778}&{1167}&{1556}&{126}&{1.23} \\
   {Minié-Danniyé}&{6}&{15}&{30}&{45}&{60}&{140}&{352}&{704}&{1057}&{1409}&{99}&{1.42} \\
   {Koura}&{8}&{20}&{41}&{62}&{83}&{120}&{301}&{603}&{904}&{1206}&{49}&{2.46} \\
   {Zgharta}&{7}&{18}&{36}&{55}&{73}&{107}&{269}&{539}&{809}&{1079}&{45}&{2.40} \\
   {Batroun}&{5}&{14}&{29}&{43}&{58}&{104}&{261}&{523}&{784}&{1046}&{75}&{1.39} \\
   {Tripoli}&{13}&{34}&{69}&{104}&{\bf 138}&{103}&{258}&{516}&{774}&{1032}&{\bf 4}&{\bf 25.80} \\
   {Marjayoun}&{6}&{17}&{34}&{51}&{68}&{101}&{254}&{508}&{762}&{1017}&{74}&{1.37} \\
   {Hermel}&{5}&{13}&{27}&{40}&{\bf 54}&{95}&{239}&{479}&{719}&{959}&{179}&{0.54} \\
   {Rachaiya}&{7}&{19}&{39}&{59}&{78}&{95}&{238}&{477}&{716}&{954}&{184}&{\bf 0.52} \\
   {Jezzine}&{6}&{15}&{31}&{47}&{63}&{62}&{155}&{311}&{467}&{623}&{64}&{0.97} \\
   {Hasbaiya}&{8}&{20}&{40}&{60}&{80}&{53}&{133}&{266}&{399}&{532}&{78}&{\bf 0.68} \\
   {Bcharré}&{5}&{13}&{27}&{41}&{55}&{35}&{88}&{176}&{265}&{353}&{41}&{0.86} \\ \bottomrule

\caption{Average Solar rooftop Potential values ($ASP$ in $MWh/year$) and Total Solar rooftop Potential values ($TSP$ in $GWh/year$) per district while varying utilization factor between 10\% and 100\%. \hl{Last column shows relative percentage of Total solar rooftop potential (at $U$ = 100\% ) with respect to the maximum hypothetical capacity ($MWC$ in $TWh/year$) for each district.} The table is sorted using $TSP$ column in descending order.}
\label{tb6}
\end{longtable}
\end{center}

\clearpage
\twocolumn

\end{document}